\documentclass{article}

\usepackage{arxiv_custom}

\usepackage[utf8]{inputenc}
\usepackage[T1]{fontenc}

\usepackage{amsmath,amssymb,amsfonts}
\usepackage{nicefrac} 
\usepackage{latexsym}
\usepackage{booktabs}
\usepackage{tablefootnote}

\usepackage{url}
\usepackage{xcolor}
\usepackage{placeins}
\usepackage{graphicx}
\usepackage{caption}
\usepackage{subcaption}

\usepackage{hyperref}

\newcommand{\he}{H\&E}
\definecolor{newcolor}{rgb}{.8,.349,.1}

\newcommand\Tstrut{\rule{0pt}{2ex}}         
\newcommand\Bstrut{\rule[-0.9ex]{0pt}{0pt}}   

\title{Leveraging weak complementary labels to improve semantic segmentation of hepatocellular carcinoma and cholangiocarcinoma in H\&E-stained slides}

\author{ 
    \hspace{1mm}Miriam Hägele$^{1,2,3}$\thanks{Contributed equally to this work.} \\
 \And
    \hspace{1mm}Johannes Eschrich$^{4,5*}$\\
 \And
    \hspace{1mm}Lukas Ruff$^3$\\
 \And
    \hspace{1mm}Maximilian Alber$^{3,6}$\\
 \And
    \hspace{1mm}Simon Schallenberg$^6$ \\
 \And
    \hspace{1mm}Adrien Guillot$^4$ \\
 \And
    \hspace{1mm}Christoph Roderburg$^7$ \\
 \And
    \hspace{1mm}Frank Tacke$^4$\\
 \And
    \hspace{2.5cm}Frederick Klauschen$^{2,5,6,8,9,+}$\\
}
\affiliations{
    \textsuperscript{\rm 1} Machine Learning Group, Technische Universität Berlin
    \textsuperscript{\rm 2} BIFOLD – Berlin Institute for the Foundations of Learning and Data 
    \textsuperscript{\rm 3} Aignostics GmbH 
    \textsuperscript{\rm 4} Department of Hepatology and Gastroenterology, Charité Universitätsmedizin Berlin 
    \textsuperscript{\rm 5} Berlin Institute of Health at Charité – Universitätsmedizin Berlin 
    \textsuperscript{\rm 6} Institute of Pathology, Charité Universitätsmedizin Berlin 
    \textsuperscript{\rm 7} Clinic for Gastroenterology, Hepatology and Infectious Diseases, University Hospital Düsseldorf 
    \textsuperscript{\rm 8} Institute of Pathology, Ludwig-Maximilians-Universität München 
    \textsuperscript{\rm 9} German Cancer Consortium, Munich Partner Site, German Cancer Research Center \vspace{0.1in} \newline 
    \textsuperscript{\rm +} Corresponding author: f.klauschen@lmu.de
}

\hypersetup{
pdftitle={Leveraging weak complementary labels to improve semantic segmentation of hepatocellular carcinoma and cholangiocarcinoma in H\&E-stained slides},
pdfauthor={Miriam Haegele et al.},
pdfkeywords={Semantic segmentation, Complementary labels, Computational pathology},
}

\begin{document}
\flushbottom
\maketitle\date{}

\begin{abstract}
 In this paper, we present a deep learning segmentation approach to classify and quantify the two most prevalent primary liver cancers -- hepatocellular carcinoma and intrahepatic cholangiocarcinoma -- from hematoxylin and eosin (H\&E) stained whole slide images. While semantic segmentation of medical images typically requires costly pixel-level annotations by domain experts, there often exists additional information which is routinely obtained in clinical diagnostics but rarely utilized for model training. We propose to leverage such weak information from patient diagnoses by deriving complementary labels that indicate to which class a sample \emph{cannot} belong to. To integrate these labels, we formulate a complementary loss for segmentation. Motivated by the medical application, we demonstrate for general segmentation tasks that including additional patches with solely weak complementary labels during model training can significantly improve the predictive performance and robustness of a model. On the task of diagnostic differentiation between hepatocellular carcinoma and intrahepatic cholangiocarcinoma, we achieve a balanced accuracy of 0.91 (CI 95\%: $0.86-0.95$) at case level for 165 hold-out patients. Furthermore, we also show that leveraging complementary labels improves the robustness of segmentation and increases performance at case level.
\end{abstract}

\section{Introduction}
Primary liver cancer is one of the most frequently diagnosed cancers worldwide. Among primary liver cancer, hepatocellular carcinoma (HCC) and intrahepatic cholangiocarcinoma (CCA) are the most frequent types accounting for roughly 70\% and 15\% of cases, respectively. Together they are among the most frequent cancers worldwide \cite{RN480}.
The diagnostic distinction between these entities has unique implications for prognosis and medical treatment. For example, certain treatment options that are regularly used in the treatment of CCA are to date ineffective and even potentially harmful for HCC, and vice versa. 
The most critical part for the classification of liver cancer is its histopathological evaluation which forms the basis for medical treatment decisions. But at the same time the histopathological classification of HCC and CCA can be challenging in some cases, even for experienced gastrointestinal pathologists \cite{RN481, RN483}.

While reliable case predictions on histological hematoxylin and eosin (H\&E) stained slides can already add value for the routine pathological workflow, for example by quickly reaching a decision for the necessity of ancillary tests, like immunohistochemistry, and thus saving time and resources, there is much more information hidden in the tissue composition. 
We hypothesize that part of this information can be accessed by using a semantic segmentation approach. In particular, the size of an area covered by a specific morphological tissue type might be correlated with relevant clinical parameters. Conventional parameters such as tumor diameter are already well-established prognostic factors for survival \cite{RN482}. Additionally, treatment response, for instance to immunotherapy, might depend on factors such as the area covered by certain immune cells and their distance to corresponding tumor cells, which can be assessed with a segmentation approach. Furthermore, segmentation approaches have the advantage of inherently providing the possibility for practitioners to verify class predictions via segmentation maps.

Semantic segmentation in medical imaging \cite{asgari2021deep} relies on pixel-wise annotations by domain experts. As labeling efforts can be extremely costly and time-consuming, annotations are often sparse and only capture a relatively small fraction of the available, usually heterogeneous data. Especially in digital pathology, where samples are in the range of gigapixels, often large parts of the data get neglected and only a very limited number of pixels are used for model training. However, there often exists additional information about the patient's whole slide images (WSIs) that is mostly ignored for training segmentation models. For example, for cases in the training set clinical diagnoses are already available from routine diagnostics and do not require further manual labeling efforts. Therefore we can incorporate this additional information at case level during model training. For the task of tumor segmentation, we cannot directly use the patient's diagnosis as weak label \cite{marini2022unleashing} due to the presence of normal tissue components even on tumorous slides. Due to the mutually exclusivity of the diagnoses, we can however assign the opposite diagnosis as complementary label, i.e.\ stating which class a case and therefore its corresponding patches do not belong to. This way we can derive complementary labels at pixel level for all cases in our training set, independent of manual labeling efforts.

Our contribution in this work is two-fold: First, we propose a segmentation approach to classify and quantify hepatocellular carcinoma and intrahepatic cholangiocarcinoma in H\&E-stained whole slide images. In contrast to classification, this approach provides more informative insights into the tissue composition such as the localization and quantification of the tumor. Additionally, the corresponding segmentation maps allow visual verification of class predictions by pathologists.
Second, we extend the segmentation approach by formulating a loss function that enables us to leverage weak complementary labels derived from patients' diagnoses. While our motivation is derived from the medical use case, our contribution regarding the utilization of complementary labels for segmentation tasks is general. We demonstrate that if only a limited number of annotated samples is available, segmentation performance can be significantly improved by a large margin via leveraging weak complementary labels on additional, not manually annotated patches. Such complementary labels at patch level are often available without further manual expenditure or at least require less skills and time. We extend this analysis for scenarios where one class is not available as complementary label. Finally, we demonstrate these benefits of leveraging complementary labels on our medical use case for semantic segmentation of hepatocellular carcinoma and cholangiocarcinoma in H\&E-stained slides.

\begin{figure*}[t]
    \centering
    \includegraphics[width=\textwidth]{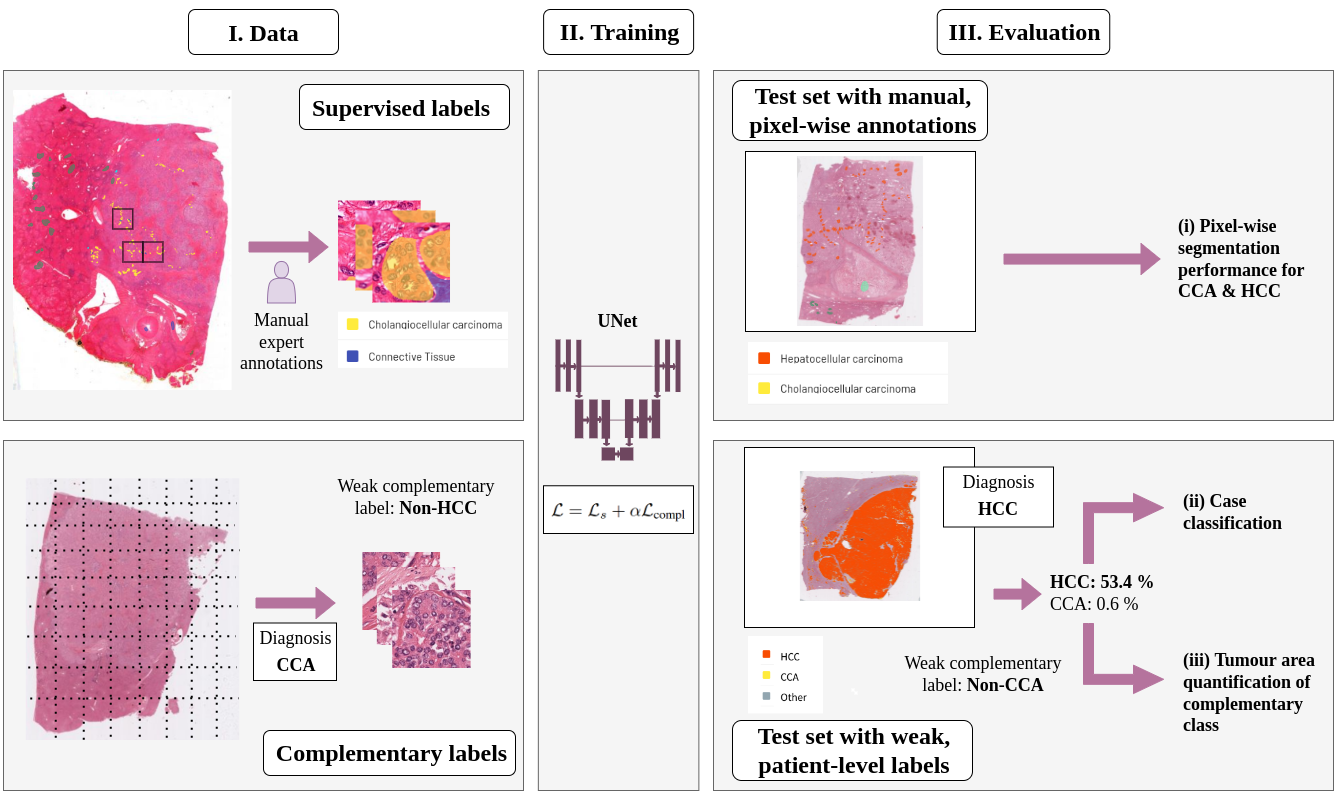}
    \caption{Mitigating the bottleneck of manual annotations by incorporating the diagnoses of patients as complementary label into the training workflow of tumor segmentation models on \he-stained whole slide images. In addition to the sparse expert annotations, complementary labels from additional whole slide images are derived from the patients diagnosis. These are incorporated into model training via a composite loss, consisting of the supervised cross-entropy part $\mathcal{L}_s$ and the complementary part $\mathcal{L}_{compl}$. For validation the segmentation models are evaluated at both the level of expert annotations as well as on measures derived from segmentation maps on unannotated data. In particular, we compute the classification performance (CCA vs. HCC) at case level based on predictions per patient which are derived from the dominance of predicted pixels. Furthermore we use the area of the complementary class as additional indication of segmentation quality.}
    \label{fig:workflow}
\end{figure*}

\section{Related work}
Regarding segmentation the U-Net \cite{ronneberger2015u} has become the de facto standard neural network architecture in medical imaging, including its successful application to tissue segmentation in histopathology (e.g.\ \cite{bulten2019epithelium, van2021hooknet, burlutskiy2018deep}).
Considering the extensive annotation efforts typically required for semantic segmentation in medical imaging, there have been several proposals in the literature to reduce this manual burden for domain experts. For example, \cite{bulten2019epithelium} use co-registered immunohistochemical (IHC) stained WSIs to extract segmentation masks. Others suggest to generate additional synthetic training patches, for example via using generative adversarial networks (GANs) \cite{bowles2018gan,hou2017unsupervised, mahmood2019deep}.\\
Another approach is to make use of complementary labels, indicating which class a sample does \emph{not} belong to, which are often easier to obtain than ordinary labels \cite{ishida2017learning, ishida2019complementary, yu2018learning}.
The use of complementary labels has recently been studied in the context of classification tasks. In particular, \cite{ishida2017learning} investigate learning solely from complementary labels. They have extended their initial approach, which had some restrictions concerning the loss function (in particular its requirement to satisfy a symmetry condition), to arbitrary models and losses \cite{ishida2019complementary}. In their approach they assume that each complementary label has the same probability of being selected given the ground truth label, that is each complementary class is sampled with a probability of $1/(k-1)$ with $k$ being the number of classes. In practice, however, complementary labels might not always be distributed uniformly across classes, for example due to selection biases of the annotators. For this reason, \cite{yu2018learning} propose a loss function that allows the probability of the complementary labels to be skewed.
To the best of our knowledge, there is only one work integrating complementary labels into semantic segmentation \cite{rezaei2020recurrent}. Whereas the focus of \cite{rezaei2020recurrent} is to explore recurrent generative adversarial models in the context of medical image segmentation, the authors additionally probe the integration of the inverse of the ground truth map as complementary labels in order to mitigate the commonly encountered class imbalance between foreground and background pixels in medical image analyses. In contrast, we aim to derive complementary labels also for pixels for which we do not have manual annotations which enables us to explore additional cases during model training.\\
Regarding machine learning-based whole slide image analyses of liver disorders, there only exist few prior works. Whereas most focus either on the classification between benign and malignant tissue in HCC (e.g.\ \cite{chen2020classification, aziz2015enhancing, huang2019automatic}) or on the classification of histological grades in HCC \cite{atupelage2014computational}, we only found one study aiming to differentiate between HCC and CCA. In particular, \cite{kiani2020impact} develop a deep learning-based assistant to help pathologists differentiate between these two most common liver cancer types on H\&E-stained whole slide images. Their approach is based on the classification of patches extracted from tumor regions which were previously annotated by a pathologist. The slide-level accuracy is subsequently reported as the mean of the patch-level probabilities.\\
Despite its great potential for medical imaging the full capability of biased complementary labels for semantic segmentation has not yet been demonstrated. With this work, we aim to fill this gap. 

\section{Methods}
In this section, we formulate a loss function for incorporating complementary labels into semantic segmentation. The proposed loss function extends the idea of biased complementary labels for classification \cite{yu2018learning} to segmentation tasks. Biased complementary labels thereby refer to labels which state an incorrect class for a pixel and where the complementary labels are not distributed uniformly across the ground truth classes.\\
The general idea of the loss is to maximize the probability of the possible ground truth classes (i.e. all classes minus the complementary class) weighted by the estimated probabilities of how likely the classes are given a particular complementary label. These probabilities need to either be estimated from a small annotated dataset or can in our case be derived from the distribution of diagnoses (for more details cf. Section~\ref{sec:model_training}). The weighted sum of the probabilities is again a probability distribution, this time across all possible ground truth classes, and can thus be optimized with standard loss functions such as the cross-entropy loss.\\
To formulate the complementary loss function formally, let $\{x_n, \bar{y}_n\}_{n=1}^N$ be the set of image patches ${x_n \in \mathbb{R}^{P \times P \times 3}}$ with corresponding complementary label masks ${\bar{y}_n \in \mathbb{R}^{P \times P}}$. Hereby, $k$ denotes the number of classes and $P$ the patch size. Assuming ${y_{n_p} \! = \! c}$ is the (unknown) ground truth label of pixel $x_{n_p}$, then the complementary label lies in ${\bar{y}_{n_p} \in \{1, ..., k\} \setminus \{c\}}$.
The estimated probabilities of assigning a complementary class $j$ given the true label $i$, i.e.\ ${Q_{ij} = P(\bar{Y}\!=\!j|Y\!=\!i)}$, are summarized in a transition matrix $Q \in \mathbb{R}^{k \times k}$. Hence, the rows of the transition matrix describe the transition probabilities of all complementary labels for the respective ground truth label. Therefore probabilities over individual rows should sum up to one. Mind that all entries on the diagonal of $Q$ will be zero, as complementary labels indicate incorrect classes. 
The benefit of capturing the probabilities of the complementary classes in such a transition matrix $Q$ is that the conditional probability of the true label $P(Y\!=\!i|X)$ can be approximated by multiplying $P(\bar{Y}\!=\!i|X)$ with the transposed transition matrix $Q^T$. Thus, we can apply standard loss functions for optimization.\\
Suppose $\hat{y}_{n_p}$ denotes the predicted softmax probabilities of pixel $x_{n_p}$ for some model, and $\bar{y}_{n_p}$ the corresponding one-hot encoded complementary label, we formulate the complementary loss as
\begin{equation}\label{eq:compl_loss}
    \mathcal{L}_{\text{compl}}(X, \bar{Y}) = - \frac{1}{NP}\sum_{n=1}^{N} \sum_{p=1}^{P} \bar{y}_{n_p} ~{\text{log}(Q^T \hat{y}_{n_p})}
\end{equation}
Note that the logarithm is applied element-wise here. Given that the transition matrix $Q$ describes all transition probabilities between complementary labels and ground truth labels, the matrix multiplication $Q^T \hat{y}_{n_p}$ consequently represents the conditional probability of the possible true labels.\\
We can further extend this loss to a focal version \cite{lin2017focal} by inserting the multiplicative factor $(1-Q^T \hat{y}_{n_p})^\gamma$ with $\gamma >0$. This penalizes the hard-to-classify pixels more strongly.\\
We then define the overall loss as the weighted sum of the supervised loss and the complementary loss,
\begin{equation}
    \mathcal{L} = \mathcal{L}_s + \alpha \, \mathcal{L}_{\text{compl}}
\end{equation}
where $\mathcal{L}_s$ denotes the categorical cross-entropy loss on the annotated pixels. Note that in contrast to the cross-entropy loss, the complementary loss is not masked and can be applied to both sparsely annotated as well as completely unannotated patches.

\subsection{Proof of concept: MNIST ablation study}
\begin{figure}
    \centering
    \includegraphics[width=0.5\textwidth]{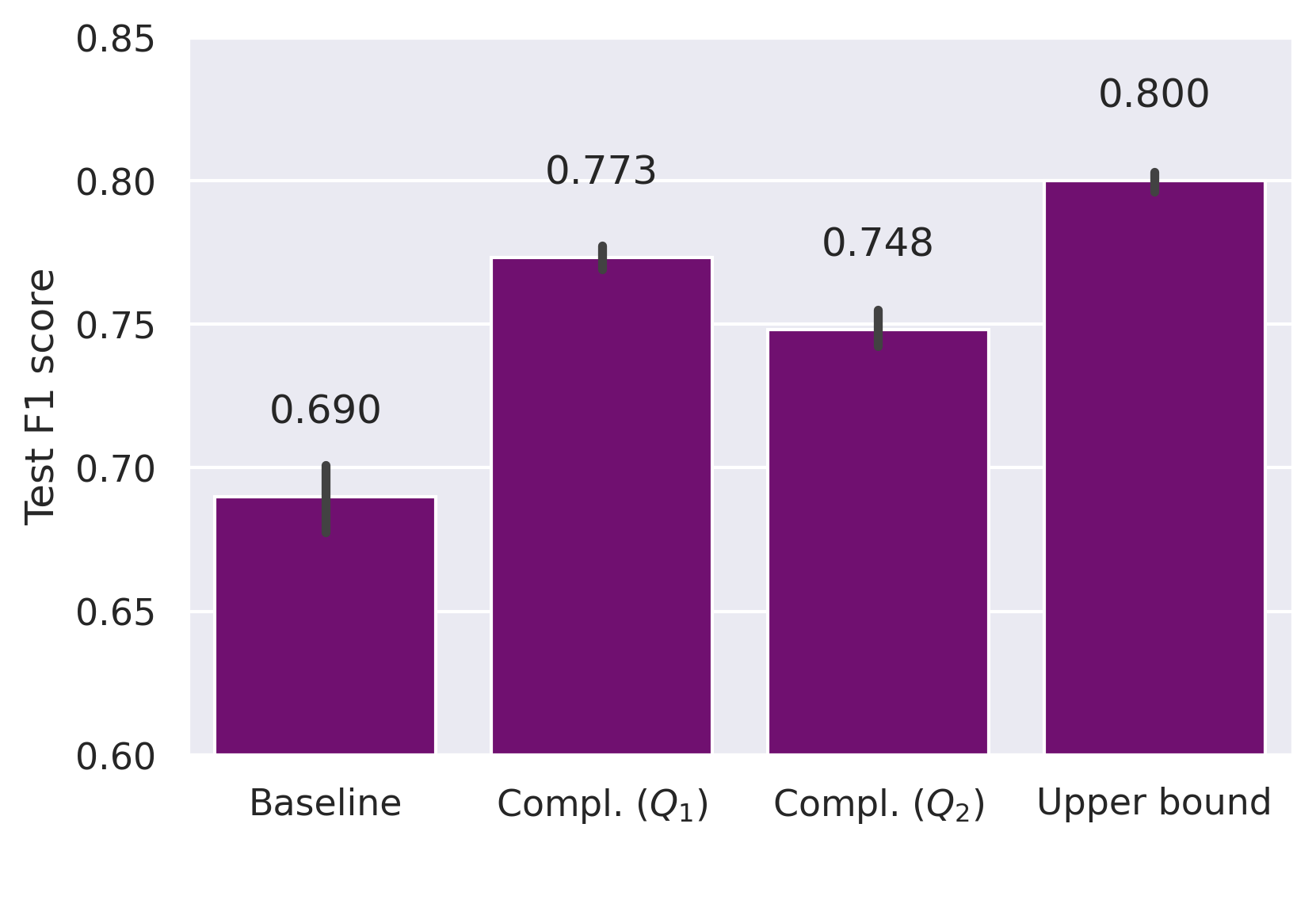}
    \caption{Proof of concept on MNIST. Only a small set of the dataset is labeled whereas the rest of the samples only contain (biased) complementary labels. Leveraging additional samples with solely complementary labels improves segmentation performance over the baseline of supervised training on the small labeled dataset. Transition matrices $Q_1$ and $Q_2$ correspond to two different underlying conditional distributions of the complementary labels. The upper bound of supervised training on the full dataset, assuming we are given supervised labels for all samples, is depicted on the right.}
    \label{fig:mnist}
\end{figure}

Here, we use the well-studied MNIST dataset to investigate the behavior of the proposed complementary loss under controlled conditions. In particular, we explore different conditional probability distributions of the complementary labels. To this end we use a subset (N=1,000) of the MNIST dataset to segment and classify the digits ``3'', ``4'' and all others. Only 10\% of the data contain supervised labels. Complementary labels for all samples were distributed according to two different conditional probability distributions where in (i) the complementary labels are biased towards respective classes and in (ii) there is no complementary label information for one of the classes. The two scenarios can formally be expressed by the two following transition matrices:
\begin{equation*}
\text{(i)} \hspace{0.3cm}
Q_1 = 
\begin{pmatrix}
  0 & .7 & .3\\
  .3 & 0 & .7 \\
  .7 & .3 & 0
\end{pmatrix}; \hspace{0.3cm}
\text{(ii)} \hspace{0.3cm}
Q_2 = 
\begin{pmatrix}
  0  & 1. & 0\\
  1. & 0  & 0 \\
  .5 & .5 & 0
\end{pmatrix}
\end{equation*}
For segmenting the digits, we train a small U-Net model and report performance over five random seeds per condition. The results shown in Fig.~\ref{fig:mnist} demonstrate that including complementary labels from additional samples in the form of the suggested complementary loss \eqref{eq:compl_loss} significantly increases performance over the supervised baseline trained on the small labeled dataset (10\% of the dataset). The expected upper bound is given by a supervised model trained on the complete dataset, thus assuming we would have access to the ground truth labels for all data points. Regarding the different distributions of complementary labels, restricting the full number of complementary classes as with $Q_2$ has a slight negative effect on performance as expected, though only slightly. Overall, we can see that utilizing complementary labels together with only 10\% supervised labels already comes close to the upper performance bound of using completely supervised labels, thus demonstrating the benefit of our proposed complementary loss for segmentation.

\section{Experiments}
In this section we will outline the segmentation approach to differentiate and quantify HCC and CCA as the primary tumor types in liver specimens, as well as applying the proposed complementary loss in this real-world scenario. In the first part of the section we will describe the machine learning-based experimental setup before introducing the dataset, which was digitized and curated for the purpose of this work. The rest of the section delineates the specific parameters and details of the model training for reproducibility.

\subsection{Experimental setup}
\label{sec:experimental_setup}
We aim to segment a given whole slide image into the following three tissue types: CCA, HCC, and non-carcinoma tissue (hereafter referred to as \textit{Other}) which contained both annotations from healthy tissue (e.g.\ normal liver epithelium, lymphocytes) as well as image artifacts. Furthermore, we study the benefits of the proposed complementary loss which allows to include more patients into model training without further manual annotation expenditure.\\
We evaluate our approach with respect to three different criteria: (i) Pixel-wise segmentation performance on the annotated test set, (ii) binary classification at case level, and (iii) quantitative evaluation of the segmentation maps on the hold-out test cohort. While evaluation against the manual annotations on a test set is the standard approach to estimate the model's generalization to unseen data, additional evaluation on not manually annotated WSIs allows for an evaluation on a much larger cohort and therefore covering more of the naturally occurring heterogeneity of hepatic liver tissue. For these cases only weak labels at case level (i.e.\ the patients' diagnoses) were available. For (iii) we use the segmentation maps to report the pixel share of the complementary class as an additional pixel-wise measure which evaluates the entirety of the whole slide image. An overview over the experimental setup is shown in Fig.~\ref{fig:workflow}.

\subsection{Data}
\label{sec:data}
\begin{figure}
    \centering
    \begin{subfigure}[b]{0.5\textwidth}
        \centering
        \includegraphics[width=\textwidth]{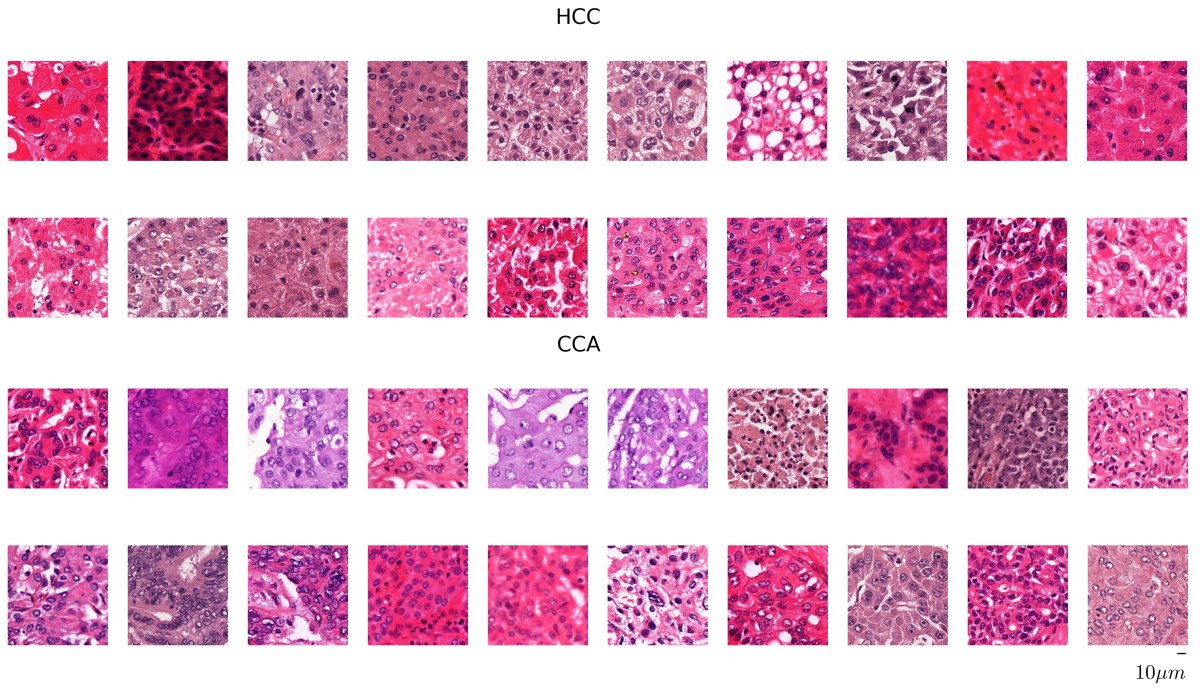}
        \caption{Tumorous patches extracted based on human annotations.}
        \label{fig:patches}
    \end{subfigure}
    \hfill
    \begin{subfigure}[b]{0.49\textwidth}
         \centering
         \includegraphics[width=\textwidth]{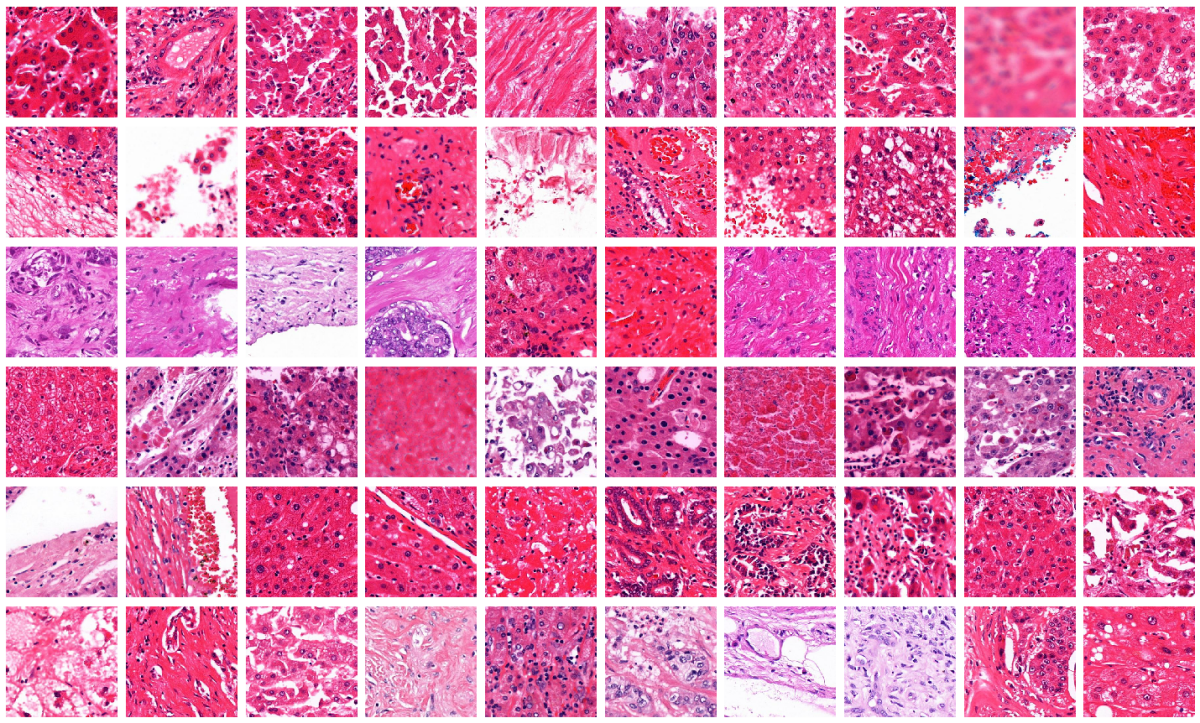}
         \caption{Patches extracted on a reagular grid from unannotated cases.}
         \label{fig:patches_compl}
    \end{subfigure}
    \caption{Examples of patches of tumorous regions of hepatocellular carcinoma (HCC) and intrahepatic cholangiocarcinoma (CCA) (left) and extracted on a regular grid from unannoted cases (right). The latter are leveraged for model training by deriving complementary labels from the patients' diagnoses. 
    The examples illustrate the high heterogeneity in morphology and staining across the dataset and thus, the associated difficulty of diagnostic differentiation.
    The patches are extracted from the corresponding whole slide images with a resolution of \mbox{0.5$\mu m$ /pixel}.}
\end{figure}

We conduct our experimental evaluation using digital whole slide images of H\&E-stained slides from formalin-fixed, paraffin-embedded (FFPE) primary hepatic tumor resections of either HCC or CCA. For this, anonymized archival tissue samples were retrieved from the tissue bank of Charité Universitätsmedizin Berlin. All data were collected in accordance with the Declaration of Helsinki and the International Ethical Guidelines for Biomedical Research Involving Human Subjects. We included tissue samples from adult patients (aged 18 and older) between 2016 and 2018 for HCC and between 2010 and 2019 for CCA, resulting in a total of 262 patients (124 CCA, 138 HCC). The histopathological classification was derived by first evaluating the morphological features in H\&E stainings. In case of diagnostic uncertainty -- e.g.\ HCC vs. CCA, HCC vs. healthy liver parenchyma, CCA vs. healthy bile duct -- additional analyses like gomori reticulin staining or immunohistochemical stainings (e.g.\ CK7, CK19, HepPar1, Glypican 3) were used.\\
Two pathologists annotated the digitized histological slides (N=124) from 97 patients (47 CCA, 50 HCC) according to the respective carcinoma and other tissue components such as healthy liver parenchyma, healthy bile duct epithelium, connective tissue, also covering commonly occurring artifacts.  The majority of annotations were collected based on corrections of segmentation maps of previous preliminary models. Therefore annotations focus on difficult regions of the slides. From the polygon annotations we extracted patches of size ${340\! \times\! 340 \, \text{px}}$ at a resolution of $0.5\mu m$, which resulted in a total of 44,088 patches. Exemplary patches for both tumor entities are shown in Fig.~\ref{fig:patches}.\\
For the additional complementary data, we derive complementary labels from the patients diagnoses. For example, if a patient is diagnosed with CCA, no patch of the patient's WSI should contain HCC and vice versa. We use the complementary label for both the sparsely annotated patches as well as patches from additional 49 unannotated patients (stratified according to diagnosis and tumor grade). These additional patches on unannotated slides were extracted on a regular grid with a stride of ten patch lengths. In total this resulted in an additional 6,143 patches. Example patches are depicted in Fig.~\ref{fig:patches_compl}.\\ 
The 165 patients (77 CCA, 88 HCC) for which we did not gather annotations, were kept as hold-out test cohort for the evaluation at case level. The distribution of tumor grades of this test cohort is provided in Tab.~\ref{tab:bac_per_grade}.

\subsection{Model training}
\label{sec:model_training}
In order to segment the different tissue types, we train a \mbox{U-Net} \cite{ronneberger2015u} with a ResNet18 backbone\footnote{Implementation taken from \url{github.com/qubvel/segmentation_models}}. The cross-entropy loss is optimized using Adam with weight decay regularization \cite{loshchilov2018decoupled} on mini-batches of 128 patches. The learning rate is experimentally chosen to be $1e-05$ and the weight decay set to $1e-05$. For the supervised part, the cross-entropy loss is class-weighted and masked (similar to \cite{bokhorst2018learning}) which is necessary due to the sparsity of annotated pixels on the patches. 
To prevent overfitting, early stopping is performed on the averaged per class $F_1$-score with a patience of 50 epochs.\\
Besides common geometric augmentations such as translation and rotation, we address the large stain color heterogeneity in the dataset (cf.\ Fig.~\ref{fig:patches}) by augmentations in the $L\alpha\beta$ color space\cite{ruderman1998statistics}. 
The advantage of such perceptual color spaces is that euclidean distances in this space are perceptually perceived as equally distant by humans. Inspired by color normalization of \cite{reinhard2001color}, we use the mean and standard deviation in the $L\alpha\beta$ color space to translate and scale the color values respectively. During training, we normalize the patch with the corresponding cases' mean and standard deviation per axis respectively, before transforming it with values randomly drawn from the fitted Gaussian distributions over the data.\\
In order to include patches from cases which were not manually annotated, the complementary labels are derived from the mutually exclusive diagnoses. While for classes \textit{HCC} and \textit{CCA} this could be computed analytically, we estimated the probabilities of complementary labels for the ground truth class \textit{Other} from the distribution of patches of both tumor types. The underlying assumption is that the share of patches displaying healthy tissue is the same for CCA and HCC cases.
Inspired by the results on the MNIST dataset (cf.\ Fig.~\ref{fig:mnist}), we additionally gathered few complementary labels for class \textit{Other}. This was achieved by assigning patches, which according to the annotation were fully covered with tumor cells, the complementary label \textit{Non-Other}. This affected about 2\% of the annotated patches. From this we can derive the following transition matrix
\begin{equation*}
Q = 
\begin{pmatrix}
  0 & .998 & .002\\
  .980 & 0 & .020 \\
  .430 & .570 & 0
\end{pmatrix}
\end{equation*}
The additional hyperparameters such as the weight of the complementary loss and the focal loss parameter were determined experimentally and were set to $\alpha=0.3$ and $\gamma=2$ throughout all experiments.

\section{Results}
We evaluate our models at pixel level on the annotated test set as well as at case level on a larger, not manually annotated cohort.
For evaluation at pixel level, results are reported as average over an outer 5-fold cross-validation, therefore taking into consideration the large heterogeneity of the dataset. This way each patient is contained in the test set once.
Both the evaluation at case level as well as a quantification of segmentation maps in terms of complementary class shares are performed on the unannotated, hold-out test cohort. To this end, we compute segmentation maps and derive their case-level prediction from the most dominant class in the segmentation map. If cases contain multiple slides, their predicted class shares are aggregated for the case prediction.

\begin{figure}
    \centering
    \includegraphics[width=0.7\textwidth]{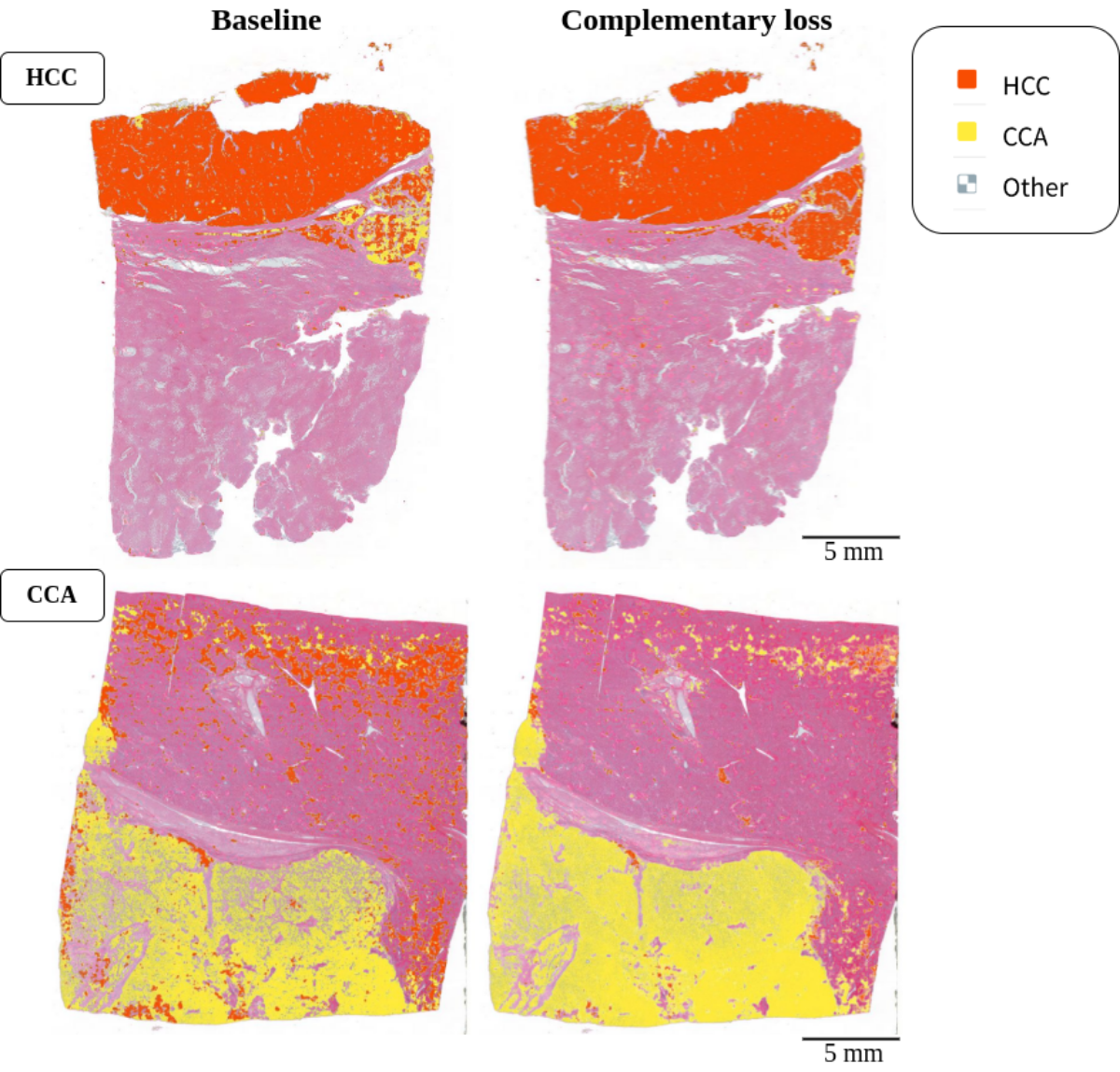}
    \caption{Segmentation maps illustrating the benefits of including the complementary loss. The left column shows the segmentation map of a baseline model while the right column shows the segmentation map of corresponding model trained with the complementary loss. Note that the prediction of class \textit{Other} is rendered transparent in these heatmaps to keep the focus on the respective tumorous regions.}
    \label{fig:heatmap_improvements}
\end{figure}

\subsection{Evaluation on the annotated test set}
\label{sec:eval_annot_test}
Segmentation performance is evaluated on the annotated, hold-out test set in order to assess the generalization capability to unseen patients. From a more practical point of view, the segmentation performance can also be interpreted as an estimate of the segmentation map quality at the granular level of cell groups. Naturally, large annotations and patients with numerous annotations will dominate pixel-level scores. To mitigate this impact, we compute a case-averaged $F_1$-score for CCA and HCC which is achieved by determining the respective $F_1$-scores per patient before averaging. This way the metric better represents the generalization to new patients instead of new annotated pixels. The performance over the outer five-fold cross-validation is ${80.20 \pm 7.07}$ for CCA and ${68.97 \pm 7.07}$ for HCC.
Out of these five models, the model which achieves median performance is used for further evaluation on the large, unannotated test set. This model achieves an $F_1$-score of $78.5$ for CCA and $69.6$ for HCC.\\
With this standard segmentation evaluation, we specifically target the performance at pixel level which is depend on the quality and representativity of the respective annotations. Therefore the scores are heavily influenced by the process of how annotations where gathered. The corrective fashion way of labeling therefore leads to annotations which are particularly focused on difficult regions of the slides (cf. Section~\ref{sec:data}).

\begin{table}
    \centering
    \begin{tabular}{p{3cm}|p{4.5cm}|p{3cm}}
    \toprule
        & \textbf{Diagnostic accuracy (95\% CI)} & \textbf{Test set size (WSI)} \Tstrut\Bstrut \\
        \hline
        Kiani et al.~\cite{kiani2020impact} & 0.885 (0.710-0.960) & 26\footnote{Validation set} \Tstrut\Bstrut \\
                                & 0.842 (0.808-0.876) & 80\footnote{Independent test set} \\
        Ours                    & \textbf{0.905} (0.861-0.947) & 165 \Tstrut\Bstrut \\
    \bottomrule
    \end{tabular}
    \newline
    \footnotesize{$^1$ Validation set, $^2$ Independent test set}
    \caption{Diagnostic accuracy of discriminating HCC and CCA.}
    \label{tab:results}
\end{table}

\begin{figure}
    \centering
    \includegraphics[width=0.3\textwidth]{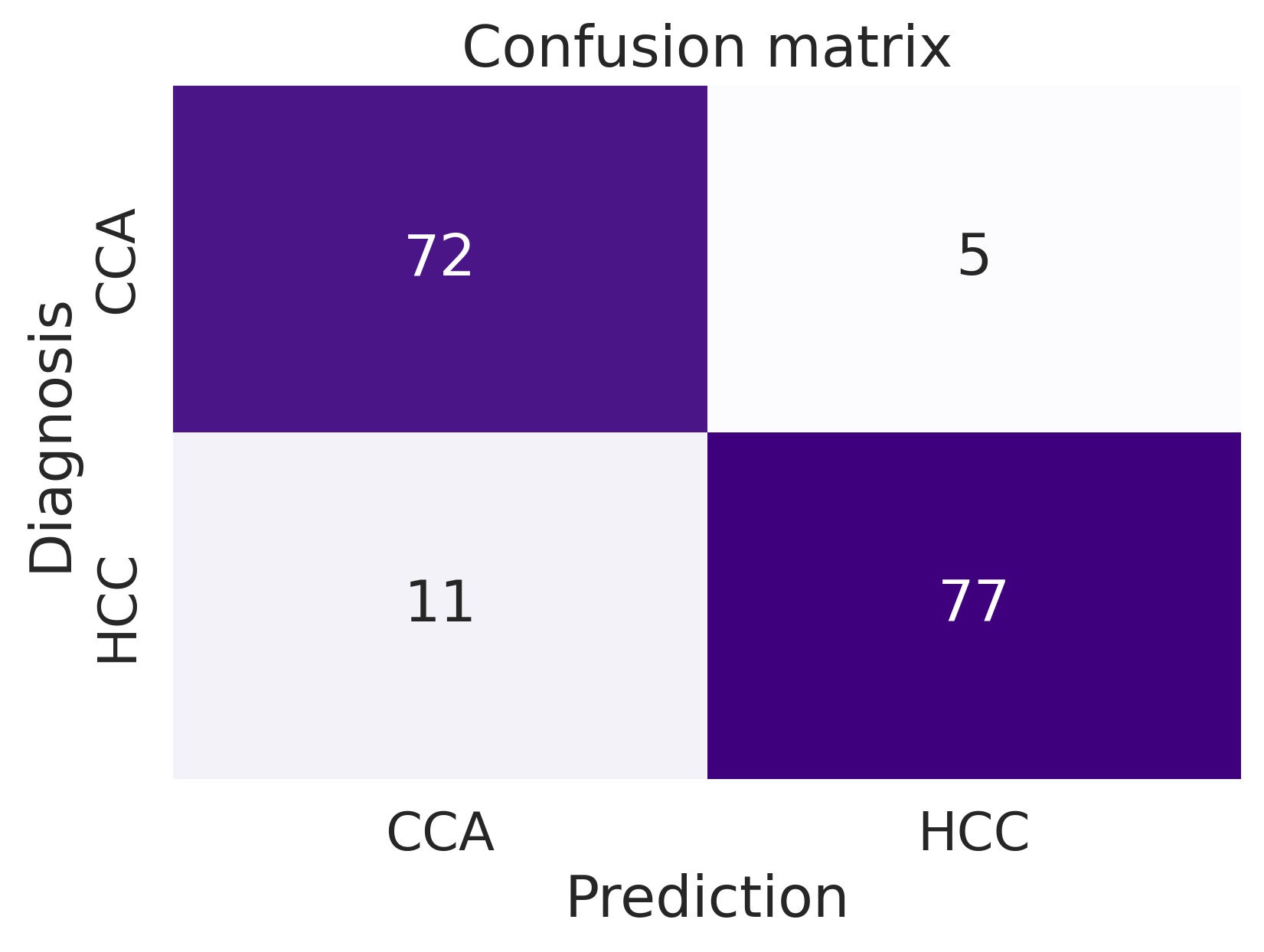}
    \caption{Confusion matrix of derived case-level predictions from the corresponding segmentation maps for the hold-out test set. The prediction at case level is determined by the dominance of either CCA or HCC pixels.}
    \label{fig:case_level_conf_matrix}
\end{figure}

\subsection{Evaluation on the unannotated test set}
\begin{figure}
    \centering
    \includegraphics[width=0.8\textwidth]{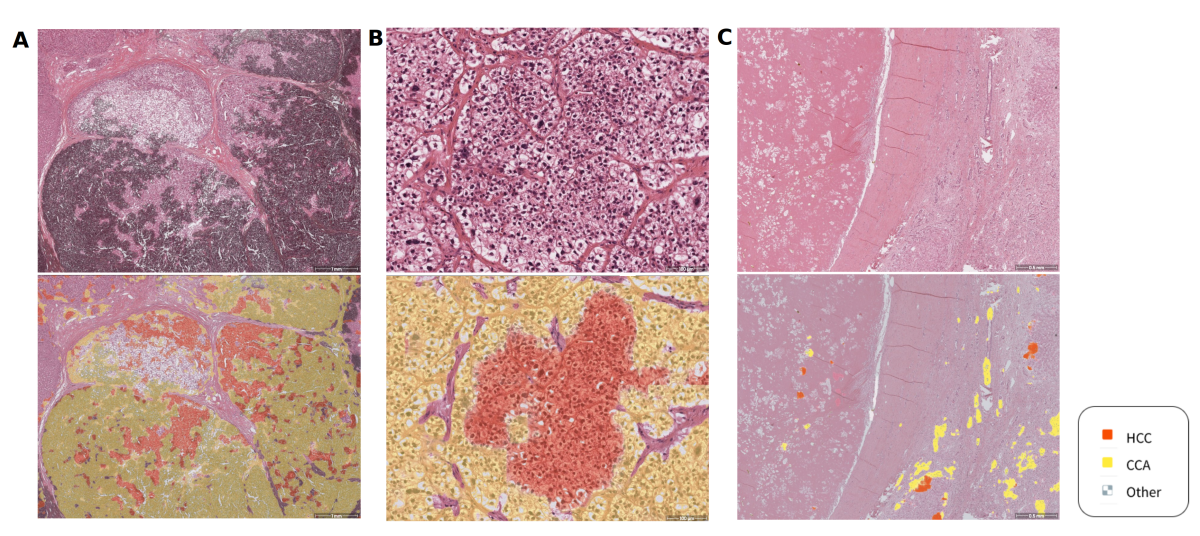}
    \caption{Examples of falsely predicted cases. A. Artefacts (black staining) lead to false predictions. Areas without black artefacts are mainly predicted correctly as either HCC or healthy tissue. B. Dedifferentiated (G3) HCC, which lost the morphological characteristics of the original healthy cells, partly falsely predicted as CCA. C. Low number of vital HCC tumor cells due to necrosis, resulting in a small HCC tumor area. Additionally, bile duct proliferations with partly dysplastic cells around the tumor had occurred, which were falsely predicted as CCA.}
    \label{fig:false_pred}
\end{figure}

In contrast, we additionally evaluate the models at case level. This is approached by deriving the case-level prediction from the predominantly predicted cancer type in the segmentation map. For instance, if the model predicts mostly HCC (in comparison to CCA), we derive the case-level label HCC. Due to the fact that this case-level evaluation is independent of manual annotations, we can assess the generalization of our model on the remaining patients for which we do not have manual annotations. For these patients only their diagnosis is available from clinical reports. As it is computationally expensive to compute gigapixel segmentation maps, we only evaluate a single model on these 165 cases. In particular, we chose the model with median performance. For case-level discrimination between HCC and CCA our model achieves a balanced accuracy of 0.905 (CI 95\%: 0.861-0.947). The reported confidence interval was obtained by bootstrapping with 1,000 resamples. Regarding the confusion between the diagnoses at case level, we observe that the model tends to confuse HCC cases as CCA. The respective confusion matrix is depicted in Fig.~\ref{fig:case_level_conf_matrix}.\\
Falsely predicted cases were reviewed by pathologists in order to identify common patterns. Several of these cases were poorly differentiated tumors, meaning they lost the morphological characteristics of the original healthy cells,  had a high percentage of artefacts or consisted of morphologically atypical tumors with mixed features in the H\&E-stainings. Moreover, in some HCC cases the tumor area was quite small (e.g. due to necrosis) and around the tumor bile duct proliferations with partly dysplastic cells had occurred, which were falsely predicted as CCA (cf. Fig.~\ref{fig:false_pred}).\\
Overall, our model outperforms previously reported results by \cite{kiani2020impact}, namely 0.885 on 26 validation WSIs and 0.842 on 80 independent test WSIs (cf. Tab.~\ref{tab:results}). It should be noted that their task was slightly easier as classification was performed on manually selected tumor regions instead of the entire whole slide image. This difference in setup is due to the main focus of \cite{kiani2020impact}, where they investigated the impact of using model predictions to assist pathologists with the subtype classification.\\
Furthermore, we use the segmentation maps to quantify the performance in terms of confusion between the cancer types at pixel level but independent of the annotations. Particularly, we measure the falsely predicted area of the complementary class separately for both carcinomas. The reported areas are relative to the slides size. This means that we compute the ratio of for example predicted CCA pixels over slide pixels for a patient diagnosed with HCC. Both carcinoma types display a similar share of complementary class area of 6.00\% and 6.15\% for CCA and HCC, respectively.

\subsection{Complementary label improvement}
\begin{figure}[tb]
    \centering
    \includegraphics[width=0.5\textwidth]{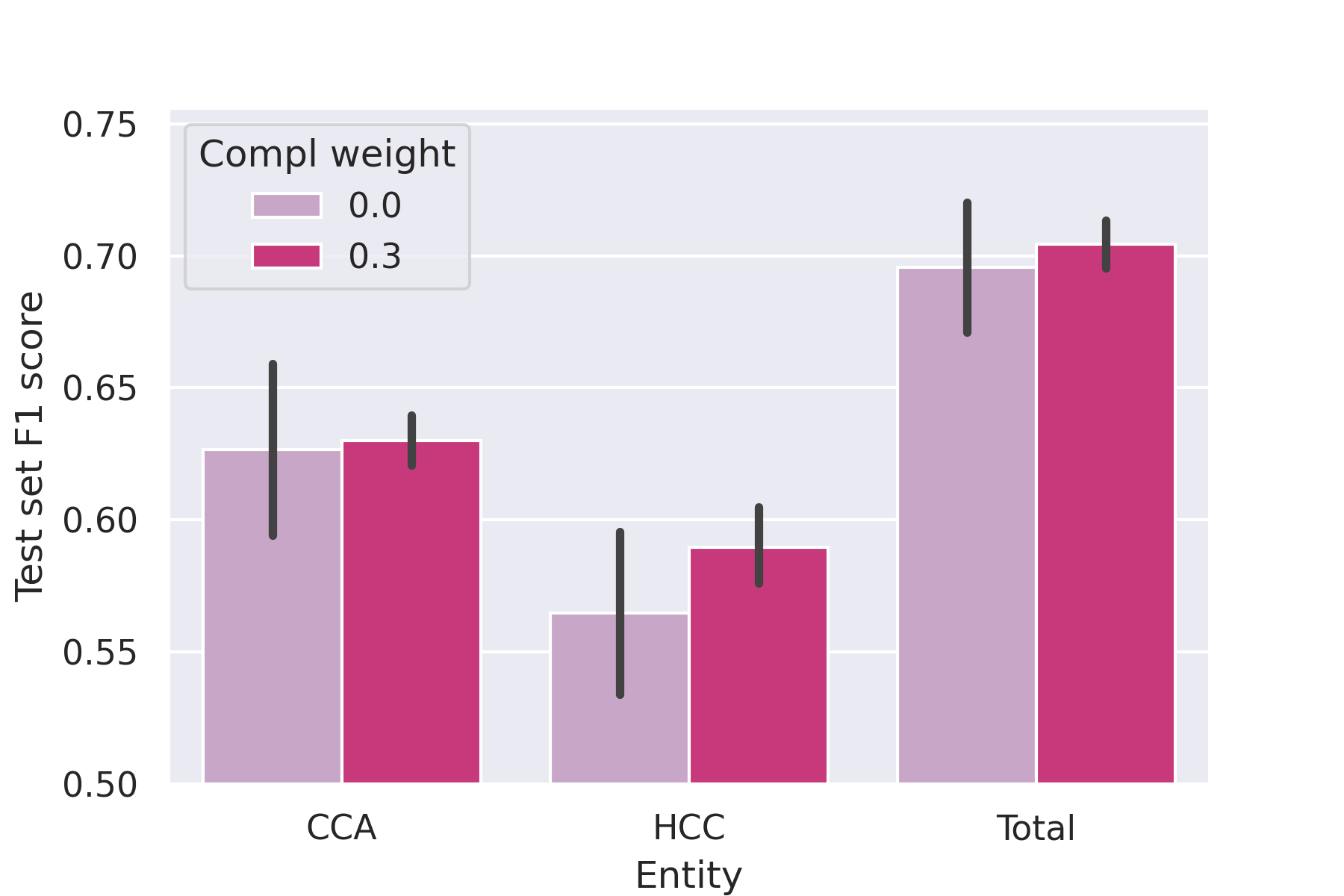}
    \caption{Test set $F_1$-scores of models trained with and without complementary labels. The plots summarize the evaluations of five models with different random seeds per condition. What stands out is the reduced variance when including complementary labels. Furthermore, while mean performance increase is small for CCA and in total, we can observe an increase in mean HCC performance.}
    \label{fig:box_f1_benefit_compl}
\end{figure}

Figure~\ref{fig:box_f1_benefit_compl} depicts the difference in segmentation performance when additionally leveraging the patients' diagnoses via the proposed complementary loss function. To not only compare approaches regarding predictive performance but also regarding robustness, performance is reported over five differently seeded runs. Due to the computational complexity of this evaluation, we evaluate the models on the (smaller) annotated test set. Besides the $F_1$-score for CCA and HCC, we also report the overall macro score, i.e.\ the average over all three segmentation classes (including \textit{Other}). Although we waived to compute case-based scores but instead directly averaged scores per class, the observed baseline trend is similar to the reported one in Sec.~\ref{sec:eval_annot_test}. By providing additional information through complementary labels to the classifier, we observe that models' prediction variance is reduced substantially for all classes. Furthermore, we observe an increase in segmentation performance, especially prominent for HCC tissue. Here, the average test set $F_1$-score over the five randomly seeded models increases by 4\%. \\
The qualitative improvement of segmentation maps when leveraging complementary labels can be seen in Fig.~\ref{fig:heatmap_improvements}. The left column shows the segmentation map of a baseline model while the right column shows the segmentation map of the corresponding model (i.e.\ using the same random seed) trained with additional complementary labels. We observe that for the HCC case, the prediction of CCA is reduced and vice versa.\\
We additionally evaluate our models at case level to assess if complementary labels also improve the balanced accuracy regarding HCC and CCA discrimination. For this reason, we compute the corresponding baselines with the same seeds. Comparing these baseline models with the models which use complementary labels, we observe an increase in case-level balanced accuracy from ${0.86 \pm 0.03}$ to ${0.91 \pm 0.03}$ on the annotated test set.\\

Overall, we observe that while the model has robust performance for both tumor types, it is more accurate in detecting CCA, both on a segmentation and case classification level. 
However, the inferior segmentation performance of HCC cases can be improved by integrating weak complementary labels (derived from the diagnoses) in terms of the proposed loss.

\section{Discussion}
The benefits of leveraging additional patches with solely weak complementary labels during segmentation model training was explored both for a general segmentation task on MNIST and on our real-world dataset. Over all experiments the additional information reduces variance and improves performance over supervised models trained on the smaller subset of annotated data. For the tumor segmentation task the performance improvement is especially observable in the mean performance of HCC tissue segmentation. Furthermore this improvement also reflects in the diagnostic classification between HCC and CCA at case level. Therefore complementary labels and the proposed loss function provide a way to include more patients without further manual annotations during training. This is especially relevant as medical segmentation datasets are often rather small while exhibiting large heterogeneity among patients. \\
We extended the idea of complementary labels to segmentation tasks which had been proven to work well in classification \cite{yu2018learning, ishida2017learning, ishida2019complementary}. Due to different properties of classification and segmentation tasks regarding low-density regions along class boundaries, some assumptions might be violated when transferring losses across these tasks. For example, the generalization of the very promising consistency regularization technique for classification \cite{laine2016temporal} was hampered by the violation of the cluster assumption in input space for segmentation tasks \cite{ouali2020semi}. Besides proving the benefits of the complementary loss in segmentation tasks, we additionally explore the situation where not all class labels are used as complementary labels and thus some classes do not have any complementary label information. While the performance increase persists, it is reduced compared to using the full range of classes as complementary labels. For the tumor segmentation task, we almost exclusively have binary complementary labels for a three class segmentation task. This means that we hardly (only for 2\% of the data) have access to patches with complementary label \textit{Non-Other}.\\
We further hypothesize that the regular grid used to extract the patches of the not annotated whole slide images might not capture the most informative structures and include some redundancy and could be improved by more sophisticated sampling strategies.
\begin{table*}
    \centering
    \begin{tabular}{p{4cm}|p{1cm}|p{1cm}|p{1cm}|p{1cm}|p{1cm}|p{1cm}|p{1cm}|p{1cm}}
    \toprule
        Tumor grade per entity  & \multicolumn{4}{c|}{\textbf{CCA}} & \multicolumn{4}{c}{\textbf{HCC}} \Tstrut\Bstrut\\
         & G1 & G2 & G3 & n/a & G1 & G2 & G3 & n/a \Tstrut\Bstrut\\
        \hline
        Balanced accuracy & \textcolor{gray}{1.0} & 0.93 & 0.92 & \textcolor{gray}{1.0} & 0.89 & 0.93 & 0.72 & \textcolor{gray}{0.0} \Tstrut\Bstrut\\
        Number of cases & \textcolor{gray}{2} & 61 & 12 & \textcolor{gray}{2} & 9 & 60 & 18 & \textcolor{gray}{1} \Tstrut\Bstrut\\
    \bottomrule
    \end{tabular}
    \caption{Balanced accuracy computed per tumor grade subgroups. \textit{n/a} hereby indicates that the grading could not be determined e.g.\ due to a lack of vital cells. Results depicted in gray are reported for completeness but we restrain from drawing conclusions as samples sizes are too small.}
    \label{tab:bac_per_grade}
\end{table*}

To better understand the performance and limitations of our tumor model, we analyzed specific subgroups with respect to tumor cell grading. 
Histological grading is a measure of the cell appearance in tumors, for liver tumors ranging from well differentiated (G1) over moderately differentiated (G2) to poorly differentiated (G3). In poorly differentiated tumors, cells lose their morphological characteristics, thus making it very difficult for pathologists to distinguish these liver tumors in H\&E-staining.  
We observe that this also translates into the performance of our model (cf. Tab.~\ref{tab:bac_per_grade}). Whereas our model achieves a balanced accuracy of 0.93 for G2 HCC cases, the performance drops to 0.72 for poorly differentiated G3 HCC cases. In CCA this difference is not so pronounced, which is in line with clinical observations, since the morphology of CCA G3 is more similar to lower grade CCAs, than this is the case for HCC.\\
A similar challenge to the above, but to a lesser extent, is that e.g.\ well differentiated tumor cells (G1) are difficult to distinguish from healthy cells, especially when only higher magnifications are used. Against this background, pathologist usually use different zoom levels in their clinical routine. While single scale at $0.5 \mu m$ per pixel seems to be sufficient for good case-level predictions, segmentation of tissue areas is in some cases challenging and could only be correctly classified by combining different zoom levels. While this was expected by pathologists, the heatmaps of G1 tumors (despite having very low numbers) show good segmentation results, which might give a hint about patterns in higher magnification, which can be used for segmentation. Nonetheless, an approach which combines various zoom levels would likely improve the model performance further.\\
While the current model shows robust performance in discriminating HCC and CCA, it is left for future work to include rare primary forms such as angiosarcomas and secondary forms of liver cancer, i.e.\ metastases from other cancers, in order to make it applicable in clinical practice.
However, our segmentation approach has scientific and potential clinical value as it allows correlation of segmentation data with clinical data. This could enable personalized diagnostic and therapeutic pathways, e.g.\ by predicting response to specific treatment options depending on the tissue composition. Follow-up projects in this regard are already underway.

\section{Conclusion}
We successfully applied a deep learning segmentation approach for diagnostic differentiation between hepatocellular carcinoma (HCC) and intrahepatic cholangiocarcinoma (CCA). Our model achieved a balanced accuracy of 0.91 at case level. 
In order to alleviate the burden of manual, time-consuming segmentation annotations by domain experts, we proposed to leverage available information from patient's diagnoses during model training. We incorporate this weak information via complementary labels, indicating that if a patient was diagnoses with HCC there should not be a prediction for CCA for this patient and vice versa. For this we formulate a complementary loss function for semantic segmentation. We provide evidence that leveraging additional patches with solely weak, complementary labels improves predictive performance for the general segmentation task as shown under controlled conditions. Furthermore, we showed that complementary labels are even beneficial if single classes are excluded from the complementary labels. In our real-world setting, we demonstrated that by including patches from not annotated patients with regard to their complementary label during model training improves the robustness of tissue segmentation and increases performance at case level.

\FloatBarrier
\section*{Acknowledgements}
This work was supported in part by the German Ministry for Education and Research as BIFOLD - Berlin Institute for the Foundations of Learning and Data (ref.\ 01IS18025A and ref.\ 01IS18037A) and as BMBF DEEP-HCC consortium and the German Research Foundation (DFG SFB/TRR 296 and CRC1382, Project-ID 403224013). J.\ E.\ is participant in the BIH Charité Junior Digital Clinician Scientist Program funded by the Charité – Universitätsmedizin Berlin, and the Berlin Institute of Health at Charité.

\bibliographystyle{abbrv}
\bibliography{main.bbl}

\end{document}